\documentclass{article} 
\usepackage{iclr2027_conference,times}

\usepackage{amsmath,amsfonts,bm}

\def\eqref#1{equation~\ref{#1}}

\def\1{\bm{1}}

\DeclareMathAlphabet{\mathsfit}{\encodingdefault}{\sfdefault}{m}{sl}
\SetMathAlphabet{\mathsfit}{bold}{\encodingdefault}{\sfdefault}{bx}{n}

\usepackage{url}
\usepackage{graphicx}
\usepackage{booktabs}
\usepackage{multirow}
\usepackage{float}
\usepackage{placeins}
\usepackage{xcolor,colortbl,tabularx}
\usepackage{threeparttable}
\usepackage{enumitem}
\usepackage[skins,breakable]{tcolorbox}
\definecolor{stoplossgray}{gray}{0.90}
\definecolor{stoplossscenariogray}{gray}{0.975}
\makeatletter
\newcommand{\stoplossmainstyle}{%
  \fontsize{10}{12}\selectfont
  \setlength{\tabcolsep}{3.5pt}%
  \renewcommand{\arraystretch}{1.10}%
  \setlength{\aboverulesep}{0.4ex}%
  \setlength{\belowrulesep}{0.6ex}%
  \setlength{\abovecaptionskip}{0pt}%
  \setlength{\belowcaptionskip}{5pt}%
  \long\def\@makecaption##1##2{%
    \vskip\abovecaptionskip
    {\normalfont\normalsize\noindent ##1: ##2\par}%
    \vskip\belowcaptionskip}}
\makeatother

\definecolor{promptblue}{HTML}{3A6EA5}
\definecolor{promptbluebg}{HTML}{F2F7FC}
\newtcolorbox{promptbox}[1]{%
  enhanced, breakable,
  colback=promptbluebg, colframe=promptblue!75!black,
  colbacktitle=promptblue!18, coltitle=black,
  fonttitle=\bfseries, before upper={\raggedright},
  boxrule=0.8pt, arc=2mm,
  left=1.4mm, right=1.4mm, top=1mm, bottom=1mm,
  title={#1}
}
\usepackage{amsmath,amssymb}
\usepackage{hyperref}
\hypersetup{hidelinks}

\title{\ifarxiv StopLoss: Teaching Agents to Predict Failure and\\Stop Early%
       \else StopLoss: Teaching Agents to\\Predict Failure and Stop Early\fi}

\author{
{\small \textbf{Jiarong Liang}$^{1}$ \quad
\textbf{Xueguang Ma}$^{1}$ \quad
\textbf{Dongfu Jiang}$^{1}$ \quad
\textbf{Yuxuan Zhang}$^{1}$ \quad
\textbf{Ping Nie}$^{1}$ \quad
\textbf{Wenhu Chen}$^{1}$ }\\[0.5em]
$^{1}$University of Waterloo \\[0.4em]
\url{https://github.com/TIGER-AI-Lab/StopLoss}
}

\makeatletter
\newcommand{\stoplossarxivmaketitle}{%
  \vbox{\hsize\textwidth\centering
    {\LARGE\upshape\bfseries \@title\par}
    \begin{tabular}[t]{@{}c@{}}\rule{\z@}{24pt}\ignorespaces\@author\end{tabular}\par
    \vskip 0.3in minus 0.1in}}
\makeatother

\newif\ifarxiv
\arxivtrue
\ifarxiv\iclrfinalcopy\fi
\makeatletter
\ifarxiv\let\@maketitle\stoplossarxivmaketitle\fi
\makeatother
\begin{document}
\raggedbottom

\maketitle
\ifarxiv\lhead{Preprint}\fi

\begin{abstract}

Coding agents keep working on runs that will fail verification, and asking a frontier model to watch for this is expensive in its own right. We introduce StopLoss, a small external monitor that learns failure recognition from completed trajectories and their verifier outcomes. The name is inspired by the stop-loss order in finance: StopLoss ends a run early when the run is likely to fail, which saves resources. Final outcomes label earlier local windows, so a fine-tuned language model can score the most recent tool steps in a single forward pass, and a threshold calibrated on successful runs turns that score into an abort decision without changing the executor. Trained on GPT-family SWE-rebench trajectories and evaluated on SWE-bench Verified and a SWE-bench Pro subset with GPT and Claude executors, StopLoss is compared against six general-purpose judges reading identical windows. Outcome supervision lifts failure prediction from near chance to a useful signal at the first check, at one to two orders of magnitude lower cost per check than any tested judge. This translates into savings: relative to completing every task, StopLoss lowers cost per retained solve by 4.5\% on Verified while preserving 99\% of successful runs, and by up to 31\% on the costlier Pro subset, reaching 48\% with a \mbox{Qwen3-8B} backbone, and in all four settings its cost per retained solve is below that of every judge policy. Outcome-trained small monitors can make failure recognition cheap enough for early stopping to pay for itself.

\end{abstract}

\begin{figure}[htbp]
\centering
\includegraphics[width=\linewidth]{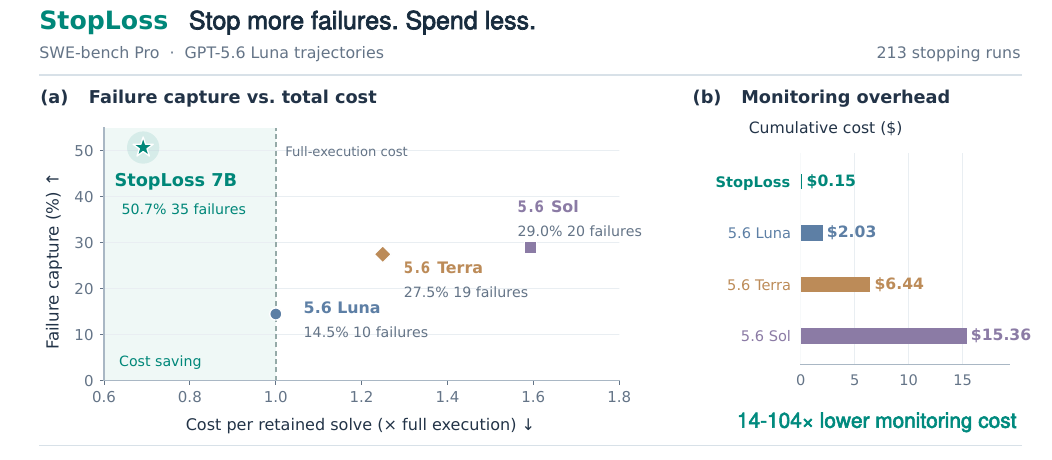}
\caption{\textbf{StopLoss stops more failing runs than LLM judges and spends far less on monitoring.}
Results on SWE-bench Pro tasks executed by \mbox{GPT-5.6} Luna.
StopLoss is the \mbox{Qwen2.5-Coder-7B-Instruct}.
(a)~Failure capture versus cost per retained solve.
StopLoss stops 35 of the 69 failing runs (50.7\%), while no judge stops more than 20.
(b)~Cumulative monitoring cost under each frozen policy: \$0.15 for StopLoss across all 213 runs, versus \$2.03--\$15.36 for the judges.
Table~\ref{tab:judges} reports details.}
\label{fig:teaser}
\end{figure}

\section{Introduction}

Coding agents can continue editing, testing, and invoking tools on runs that will ultimately fail verification. By the time the verdict arrives, the cost of those runs has already been paid. A monitor that recognizes failure risk from what is observable during execution could stop unproductive work before more computation is spent.

Two kinds of monitors are natural, and each has a problem. Budget-based approaches, from step or spend caps to the within-budget feasibility estimators of BAGEN \citep{lin2026bagen}, ask whether a run will finish within a budget, so a run that would eventually pass can still be cut off. A general-purpose model can instead read the trajectory and judge the agent's progress, but each judgment is itself an expensive call, and repeated judging can cost more than it saves. The underlying constraint is simple: early stopping lowers the cost per solved task only if the execution it avoids exceeds the monitoring overhead plus the successes it mistakenly kills (Section~\ref{sec:costdrivers}). A practical monitor must therefore be both discriminative and cheap.

Completed trajectories offer a way to obtain both. Verifier outcomes arrive only after execution, but in retrospect they label every earlier window of a run, providing supervision for a predictor of eventual failure from local records alone. Because the label is the outcome of continuing the observed execution, an expensive run that eventually passes remains a success: the monitor learns to recognize failure rather than to enforce a budget. A small model trained on such data can then assess new runs as they unfold in a single forward pass, reading only observable text without the hidden-state access required by probe-based approaches \citep{ruan2026doomed}, so it can monitor closed executors.

\emph{StopLoss} implements this idea. We adapt 7B--8B models with LoRA \citep{hu2022lora} on GPT-family SWE-rebench trajectories. At scheduled checks, the monitor scores a short window of recent actions and observations and reads a failure probability directly from the output logits. A threshold calibrated on successful trajectories maps the score to an abort decision; the executor remains fixed. We evaluate on SWE-bench Verified and a three-repository SWE-bench Pro subset with \mbox{GPT-5.6} Luna and Claude Sonnet~4.6 executors, comparing against six general-purpose judges, three per executor family, that read identical windows, and charging every method for the checks it visits and the successes it loses.

The results connect learned recognition to inexpensive monitoring and early stopping. Figure~\ref{fig:teaser} illustrates this connection on Pro/Luna. Outcome supervision moves failure ranking from near chance to a usable signal: on Verified/Luna, training raises Qwen2.5-7B-Instruct's mean single-model tier AUC from 0.491 to 0.606, and the trained ensemble already separates outcomes at the first scheduled check, tool step 5 (Section~\ref{sec:detect}). Each check costs an estimated \$0.00024, one to two orders of magnitude below every tested judge (Section~\ref{sec:paired}). On calibration-disjoint Verified/Luna runs, the frozen policy retains 99.0\% of otherwise successful runs and lowers cost per retained solve by 4.5\% relative to completing every run, whereas every GPT-judge policy raises it; savings reach 31\% on the costlier Pro subset, and across all four settings StopLoss has lower estimated cost per retained solve than every judge policy. A judge can rank failures better and still lose money: on Verified/Sonnet, Claude Opus~5 has the highest AUC in the study yet raises cost per solve by a third, because its monitoring bill exceeds the execution it saves.

Our contributions are:
\begin{itemize}[leftmargin=1.2em,topsep=2pt,partopsep=0pt,itemsep=1pt,parsep=0pt,beginpenalty=10000]
\item \textbf{Outcome-supervised failure monitoring.}
We show that completed trajectories and their verifier outcomes suffice to train a small external monitor that scores local execution windows and triggers early stopping, without modifying the executor or accessing its internals.
\item \textbf{Early recognition at low monitoring cost.}
Outcome supervision supplies failure signal at the first scheduled check, at per-check costs one to two orders of magnitude below general-purpose judges.
\item \textbf{Practical early-stopping evaluation.}
Using replay that charges monitoring and lost successes, and a break-even analysis that relates the two, we show cost advantages over six judge policies across four executor--benchmark settings and three 7B--8B backbones, and characterize when prediction and threshold transfer support early stopping.
\end{itemize}

\section{StopLoss}
\label{sec:setting}
\label{sec:data}

\begin{figure}[t]
\centering
\includegraphics[width=\linewidth]{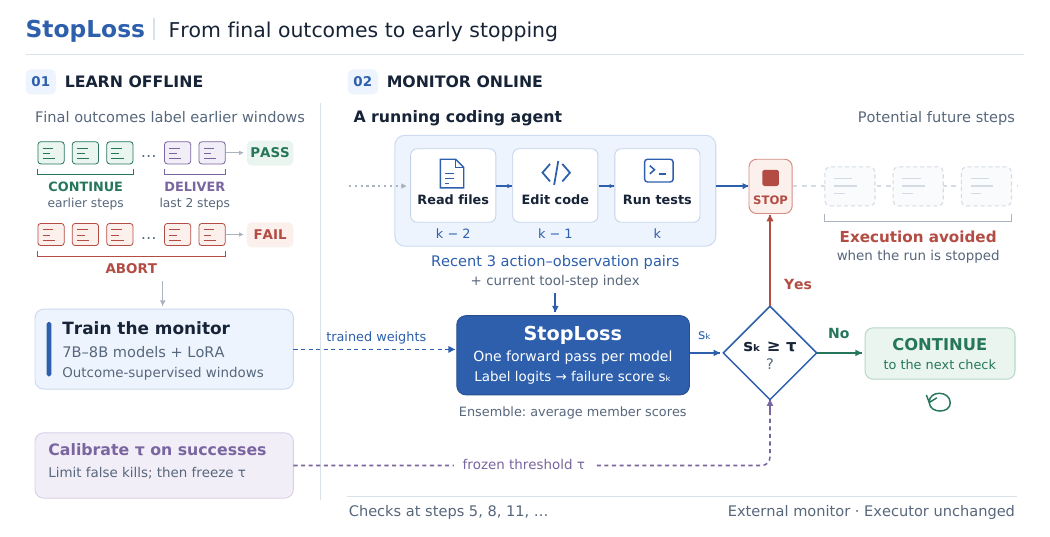}
\caption{\textbf{StopLoss turns final verifier outcomes into early-stopping decisions.}
Offline, the outcome of each completed run labels its earlier windows, 7B--8B models are adapted with LoRA on these windows, and the stopping threshold $\tau$ is calibrated on successful runs and then frozen.
\mbox{Online}, at tool steps $5,8,11,\ldots$, each ensemble member reads the step count and the three most recent action--observation pairs in one forward pass; the run stops once the averaged failure score reaches~$\tau$, and the executor itself is unchanged.}
\label{fig:workflow}
\end{figure}

StopLoss learns failure signals from completed trajectories, scores local execution windows, and stops runs when a calibrated threshold is crossed. The executor remains fixed. Figure~\ref{fig:workflow} summarizes the workflow.

\subsection{Trajectory windows and outcome supervision}
\label{sec:windows}

For task $t$, a completed trajectory has $n_t$ tool steps $\mathcal T_t=((a_{t,1},o_{t,1}),\ldots,(a_{t,n_t},o_{t,n_t}))$ and verifier outcome $y_t\in\{0,1\}$, where $y_t=1$ denotes success. At step $k$, the monitor reads the step count and the most recent three action--observation pairs as window $w_{t,k}$. We use a 1,600-token sequence limit; Appendix~\ref{app:prompt} specifies the encoding and truncation rules.

Final outcomes supervise earlier local windows: windows from failed runs receive \textsc{Abort}; those at the last two tool steps of successful runs ($k\geq n_t-1$) receive \textsc{Deliver}, and earlier successful-run windows receive \textsc{Continue}. \textsc{Deliver} is an auxiliary training label; runtime stopping uses only the \textsc{Abort} score. Outcomes supply offline labels, while runtime inputs contain only trajectory text observed by the current checkpoint.

\subsection{Learning and failure scoring}
\label{sec:scoring}

We adapt a pretrained 7B--8B causal language model with LoRA \citep{hu2022lora}, using the outcome-derived labels and the standard causal language-model objective with loss computed only on target-answer tokens. Encoding and optimization details are in Appendix~\ref{app:recipes}.

For scoring, let $z_c(w_{t,k})$ be the next-token logit at the end of the templated prompt for the first token of candidate answer $c\in\mathcal A=\{\textsc{Continue},\textsc{Deliver},\textsc{Abort}\}$. We read the failure score as
\[
 s_t(k)=\frac{\exp z_{\textsc{Abort}}(w_{t,k})}
 {\sum_{c\in\mathcal A}\exp z_c(w_{t,k})}.
\]
A check requires one forward pass per model, without generating an answer or reasoning trace. The small base, bounded input, and direct logit readout limit monitoring computation. An ensemble evaluates every member and averages their failure scores.

\subsection{Calibration and early stopping}
\label{sec:stoprule}

Calibration and deployment share checkpoints $\mathcal K_t=\{5,8,11,\ldots\}\cap[1,n_t]$. For each successful calibration run reaching a checkpoint, we compute $z_t=\max_{k\in\mathcal K_t}s_t(k)$. This maximum captures the trajectory-level false-kill event: one threshold crossing aborts the entire run. We choose $\tau$ so that at most a target fraction $\rho$ of these successes would be stopped (Appendix~\ref{app:thresholds}). This is an empirical calibration constraint, not a guarantee under distribution shift.

During execution, a checkpoint score $s_t(k)\geq\tau$ triggers an abort; otherwise the executor continues. The stopping time is
\[
 K_t(\tau)=\min\{k\in\mathcal K_t:s_t(k)\geq\tau\},
 \qquad \min\varnothing=\infty.
\]
\section{Experimental setup}
\label{sec:protocol}

\subsection{Data, executors, and evaluation cohorts}
\label{sec:cohorts}
\label{sec:calibration-setup}

Training trajectories come from SWE-rebench \citep{badertdinov2025swerebench}, collected with GPT family executors in Harbor's Terminus-2 scaffold \citep{harbor2026framework}. We fit on 4,867 tasks (61,315 windows) and reserve 219 tasks from held-out repositories (1,858 windows) for model selection. External test tasks are disjoint from both sets.

Evaluation targets SWE-bench Verified's 500 tasks \citep{jimenez2024swebench,openai2024verified} and the original public SWE-bench Pro release's 266-task Python subset \citep{deng2025swebenchpro}, spanning \texttt{ansible}, \texttt{openlibrary}, and \texttt{qutebrowser}. GPT-5.6 Luna and Claude Sonnet~4.6 both use stock Terminus-2 and benchmark-provided environments; benchmark verifiers supply outcomes. Runs lacking a valid verdict are excluded, with collection details in Appendix~\ref{app:extension}.

\paragraph{Threshold calibration.}\looseness=-1
Methods evaluating GPT-5.6 Luna trajectories use 90 successful GPT runs from the 219 SWE-rebench model-selection validation tasks, not from the 500 Verified targets. Methods evaluating Claude Sonnet~4.6 trajectories use 76 of its successful early exploratory runs (51 from Verified, 25 from Pro) to adjust thresholds without updating monitor parameters. Each method obtains its own threshold under the nominal $\rho=5\%$ false-kill target (Section~\ref{sec:stoprule}), frozen for evaluation. GPT calibration reuses model-selection tasks, so it is not an independent certification set. Appendix~\ref{app:thresholds} gives checkpoint coverage and threshold computation.

\paragraph{Ranking and stopping cohorts.}
We distinguish score ranking from evaluation of the calibrated stopping policy. Ranking uses all 1,523 valid runs on 763 tasks: 499 Verified runs per executor and 264/261 Pro runs for GPT-5.6 Luna/Claude Sonnet~4.6. It includes the Claude Sonnet~4.6 calibration runs because AUC measures score ordering without applying a threshold. To evaluate stopping on tasks unseen during calibration or early exploration, we exclude all 122 exploratory tasks (74 Verified and 48 Pro) and require valid verdicts from both executors. This leaves the same 425 Verified and 213 Pro tasks for each executor, totaling 1,276 runs on 638 tasks, disjoint from both calibration pools. Thus, Verified's 500 targets become 499 valid runs and then 425 stopping runs after removing 74 exploratory tasks; the excluded tasks are not all calibration examples. Appendix~\ref{app:coverage} gives per-setting details.

\paragraph{Cohorts used in tables and figures.}
Tables~\ref{tab:judges} and~\ref{tab:full} report AUC on the full ranking cohort; cost and stopping outcomes in Table~\ref{tab:judges} use the independent stopping cohort. Figure~\ref{fig:teaser} uses the Pro/Luna stopping cohort of Table~\ref{tab:judges}. Figure~\ref{fig:checkpoint} evaluates the original runs reaching each checkpoint before stopping is applied, with sample counts shown in the figure. The descriptive stopping-timing analysis in Section~\ref{sec:timing} uses the full Verified/Luna cohort.

\subsection{Monitors and comparison methods}
\label{sec:judge}

We train Qwen2.5-7B-Instruct, Qwen3-8B, and Qwen2.5-Coder-7B-Instruct \citep{qwen2024qwen25,qwen2025qwen3,hui2024qwen25coder} with a common LoRA recipe (Appendix~\ref{app:recipes}). We include all five trained Qwen2.5-7B-Instruct seeds and all four seeds for each other base, without test-based selection. Each ensemble pays the sum of its members' inference costs. The primary five-member Qwen2.5-7B-Instruct ensemble, used for StopLoss results unless otherwise specified, was frozen before formal judge scoring. We designated it as primary because the recipe was developed on this backbone; the other two bases test whether the recipe transfers. Untuned bases use the same readout (Section~\ref{sec:scoring}); Qwen3-8B thinking is disabled.

Judging follows the same-family protocol in calibration and evaluation: GPT-5.6 Sol, GPT-5.6 Terra, and GPT-5.6 Luna read GPT-5.6 Luna trajectories; Claude Opus~5, Claude Opus~4.6, and Claude Sonnet~4.6 read Claude Sonnet~4.6 trajectories. All small monitors read both executors. Methods receive identical local window bodies within each cell, with a probability-estimation suffix for judges. GPT judges use high reasoning effort; Claude judges use adaptive thinking with high effort. Prompts and configurations are fixed (Appendix~\ref{app:prompt}); Table~\ref{tab:prices} lists model identifiers.

The step-count-only ranking baseline uses $s_t(k)=k$, without reading trajectory content.

\subsection{Metrics and cost accounting}
\label{sec:metrics}

\paragraph{Stopping and cost.}
We evaluate the abort-only policy by replaying complete trajectories. Let $C_t(k)$ be execution cost through step $k$, with $C_t(\infty)$ including all execution calls, and $M_t(\tau)$ the cost of checks visited through stopping or completion. Cost per retained solve (CPS) is
\begin{equation}
 \mathrm{CPS}(\tau)
 =\frac{\sum_t\{C_t(K_t(\tau))+M_t(\tau)\}}
 {\sum_t y_t\,\mathbf{1}[K_t(\tau)=\infty]}.
\end{equation}
The false-kill rate (FK) is the fraction of otherwise successful runs stopped; success retention is $1-\mathrm{FK}$. Failure capture is the fraction of otherwise failed runs correctly stopped. Runs without checkpoints always survive, pay full execution cost, and incur no monitoring cost. We report $\Delta\mathrm{CPS}=100(\mathrm{CPS}/\mathrm{CPS}_0-1)$, where $\mathrm{CPS}_0$ is the cost per solve under full execution without monitoring; negative values indicate savings. Comparisons include success retention because equal nominal targets do not ensure equal realized risk.

\paragraph{Prediction.}
Failure windows often occur later and at higher spend. Our primary ranking metric, \emph{tier AUC}, divides each cell's checkpoint windows into quartiles of spent execution cost, computes failure AUC within each tier containing both outcomes, and averages valid tiers. This compares windows at similar accumulated costs, although longer runs still contribute more observations. Fixed-step diagnostics instead compute ordinary AUC with one observation per run reaching that step, before any stopping policy.

\paragraph{Cost accounting.}
We charge execution through stopping and all visited checks, including recorded judge attempts and every ensemble member. Table~\ref{tab:judges} reports cumulative monitoring cost $\sum_t M_t(\tau)$; Section~\ref{sec:paired} reports per-check costs. Under the frozen September 18 list-price schedule, cached input pays the full input rate, and output includes reasoning tokens. Small-monitor costs use batch-1 scoring-loop GPU time; fixed rules cost zero. Per-run costs exclude training, infrastructure, one-off calibration, and offline scoring of unvisited windows. Appendix~\ref{app:accounting} provides hardware, timing, prices, and call accounting; Section~\ref{sec:costdrivers} and Appendix~\ref{app:costsensitivity} analyze proxy-cost sensitivity.

\subsection{Statistical protocol and study scope}
\label{sec:statistics}

We use paired repository-cluster bootstrap draws shared across methods and executors within each benchmark \citep{efron1994bootstrap}. The Pro Python subset has only three repository clusters, so its intervals should be read with caution. Bootstrap intervals describe test-set uncertainty conditional on the trained models. Full judge coverage was commissioned after inspecting initial judge results and full small-monitor scores; intervals do not adjust for that expansion decision (Appendix~\ref{app:extension}).

\section{Main results}
\label{sec:results}

\subsection{Cost-effective early stopping}
\label{sec:deploy}

On the independent Verified/Luna stopping cohort, StopLoss combines low monitoring overhead with high success retention (see Table~\ref{tab:judges}). Its estimated cumulative monitoring cost is \$0.44, and it retains 99.0\% of otherwise successful runs, compared with 89.5--91.8\% for the three GPT-judge policies. At the frozen operating points, its monitoring-inclusive cost per retained solve is approximately 12--56\% lower than that of these policies. Compared with running all tasks to completion without monitoring, its estimated cost per retained solve is 4.5\% lower.

The larger cost reductions on Pro highlight the economic value of inexpensive monitoring across evaluation benchmarks. Despite similar tier AUCs across Verified and Pro, StopLoss achieves estimated cost-per-solve reductions of 24.2\% on Pro/Luna and 30.6\% on Pro/Sonnet. These reductions accompany success retention of 72.2\% and 91.6\%, respectively, showing that the economic benefit depends on the stopping behavior and execution costs in each setting, beyond predictive discrimination alone.

The advantage over judges is not specific to the primary backbone. In all four settings, the \mbox{Qwen3-8B} and \mbox{Qwen2.5-Coder-7B-Instruct} ensembles also have lower estimated cost per retained solve than every judge policy (see Table~\ref{tab:judges}). On Pro/Sonnet, the \mbox{Qwen3-8B} ensemble lowers cost per retained solve by 48.1\% while retaining 83.2\% of otherwise successful runs, versus 30.6\% and 91.6\% for the primary ensemble. 

\begin{table}[!t]
\centering
\stoplossmainstyle
\begin{threeparttable}
\caption{
\textbf{StopLoss achieves lower estimated cost per retained solve than LLM-judge policies across all four settings and all three backbones.}
$\Delta$CPS reports the percentage change in cost per retained solve relative to full execution without monitoring, including monitoring overhead; negative values indicate savings.
Monitoring is the estimated cumulative cost of monitor calls during execution.
Tier AUC measures how well the monitor assigns higher failure scores to runs that eventually fail than to those that succeed (0.5 = chance; 1.0 = perfect ranking).
Success retained is the percentage of otherwise successful runs preserved by the policy; failure capture is the percentage of otherwise failed runs correctly stopped.
Within each setting, bold marks the best value in each cost and stopping-outcome column.
StopLoss denotes the primary \mbox{Qwen2.5-7B-Instruct} ensemble.
}
\label{tab:judges}
\begin{tabularx}{\linewidth}{l>{\raggedleft\arraybackslash}X<{\hspace*{9.1pt}}>{\raggedleft\arraybackslash}X<{\hspace*{9.5pt}}>{\raggedleft\arraybackslash}X<{\hspace*{12.0pt}}>{\raggedleft\arraybackslash}X<{\hspace*{12.0pt}}>{\raggedleft\arraybackslash}X<{\hspace*{14.5pt}}}
\toprule
\multirow{2}{*}{\textbf{Method}} & \multicolumn{2}{c}{\textbf{Cost}} & \multicolumn{1}{c}{\textbf{Prediction}} & \multicolumn{2}{c}{\textbf{Stopping outcomes}} \\
\cmidrule(lr){2-3}\cmidrule(lr){4-4}\cmidrule(lr){5-6}
 & \multicolumn{1}{c}{\shortstack{$\Delta$CPS\\(\%) $\downarrow$}} & \multicolumn{1}{c}{\shortstack{Monitoring\\(\$) $\downarrow$}} & \multicolumn{1}{c}{\shortstack{Tier AUC\\$\uparrow$}} & \multicolumn{1}{c}{\shortstack{Success\\retained\\(\%) $\uparrow$}} & \multicolumn{1}{c}{\shortstack{Failure\\capture\\(\%) $\uparrow$}} \\
\midrule
\rowcolor{stoplossscenariogray}
\multicolumn{6}{l}{\textbf{SWE-bench Verified / GPT-5.6 Luna}\quad 425 stopping runs; 306 successes; 119 failures} \\
\rowcolor{stoplossgray}
\textbf{StopLoss} & \textminus 4.5 & 0.44 & 0.622 & \textbf{99.0} & 5.0 \\
\rowcolor{stoplossgray}
\quad Qwen3-8B & \textminus 1.8 & 0.43 & 0.605 & 97.7 & 5.0 \\
\rowcolor{stoplossgray}
\quad Qwen2.5-Coder-7B-Instruct & \textbf{\textminus 5.2} & \textbf{0.35} & 0.612 & \textbf{99.0} & 5.9 \\
GPT-5.6 Sol & +116.8 & 33.20 & 0.633 & 91.8 & 20.2 \\
GPT-5.6 Terra & +38.6 & 12.90 & 0.623 & 89.5 & \textbf{24.4} \\
GPT-5.6 Luna & +8.4 & 4.18 & 0.600 & 90.8 & 14.3 \\
\addlinespace[0.65ex]
\rowcolor{stoplossscenariogray}
\multicolumn{6}{l}{\textbf{SWE-bench Verified / Claude Sonnet 4.6}\quad 425 stopping runs; 326 successes; 99 failures} \\
\rowcolor{stoplossgray}
\textbf{StopLoss} & +0.2 & 0.44 & 0.561 & \textbf{100.0} & 0.0 \\
\rowcolor{stoplossgray}
\quad Qwen3-8B & \textbf{\textminus 1.1} & 0.42 & 0.563 & 99.7 & 2.0 \\
\rowcolor{stoplossgray}
\quad Qwen2.5-Coder-7B-Instruct & +0.1 & \textbf{0.35} & 0.561 & \textbf{100.0} & 0.0 \\
Claude Opus 5 & +34.7 & 127.13 & 0.747 & 95.7 & \textbf{32.3} \\
Claude Opus 4.6 & +11.6 & 37.07 & 0.602 & 99.1 & 0.0 \\
Claude Sonnet 4.6 & +10.3 & 37.97 & 0.602 & 97.5 & 3.0 \\
\addlinespace[0.65ex]
\rowcolor{stoplossscenariogray}
\multicolumn{6}{l}{\textbf{SWE-bench Pro / GPT-5.6 Luna}\quad 213 stopping runs; 144 successes; 69 failures} \\
\rowcolor{stoplossgray}
\textbf{StopLoss} & \textminus 24.2 & 0.20 & 0.614 & 72.2 & 43.5 \\
\rowcolor{stoplossgray}
\quad Qwen3-8B & \textminus 25.7 & 0.21 & 0.621 & 84.0 & 37.7 \\
\rowcolor{stoplossgray}
\quad Qwen2.5-Coder-7B-Instruct & \textbf{\textminus 30.9} & \textbf{0.15} & 0.620 & 74.3 & \textbf{50.7} \\
GPT-5.6 Sol & +59.3 & 15.36 & 0.657 & 86.8 & 29.0 \\
GPT-5.6 Terra & +24.9 & 6.44 & 0.651 & 79.2 & 27.5 \\
GPT-5.6 Luna & +0.0 & 2.03 & 0.639 & \textbf{87.5} & 14.5 \\
\addlinespace[0.65ex]
\rowcolor{stoplossscenariogray}
\multicolumn{6}{l}{\textbf{SWE-bench Pro / Claude Sonnet 4.6}\quad 213 stopping runs; 107 successes; 106 failures} \\
\rowcolor{stoplossgray}
\textbf{StopLoss} & \textminus 30.6 & 0.30 & 0.579 & 91.6 & 19.8 \\
\rowcolor{stoplossgray}
\quad Qwen3-8B & \textbf{\textminus 48.1} & \textbf{0.23} & 0.608 & 83.2 & \textbf{31.1} \\
\rowcolor{stoplossgray}
\quad Qwen2.5-Coder-7B-Instruct & \textminus 34.5 & \textbf{0.23} & 0.578 & 89.7 & 19.8 \\
Claude Opus 5 & \textminus 26.5 & 34.59 & 0.695 & 95.3 & \textbf{31.1} \\
Claude Opus 4.6 & \textminus 17.2 & 23.98 & 0.614 & \textbf{97.2} & 15.1 \\
Claude Sonnet 4.6 & \textminus 29.0 & 19.84 & 0.598 & 83.2 & 23.6 \\
\bottomrule
\end{tabularx}
\end{threeparttable}
\end{table}
\subsection{Learning early failure recognition}
\label{sec:detect}

\textbf{Training provides most of the prediction gain, with smaller gains from ensembling.}
For Qwen2.5-7B-Instruct, training improves mean single-model tier AUC by
$+0.115$ on Verified/Luna and $+0.148$ on Pro/Luna;
ensembling adds approximately $0.014$--$0.016$.
All three trained ensembles achieve higher AUC point estimates
than their untuned bases across all four settings
(Table~\ref{tab:full}).
Appendix~\ref{app:diagnostics} reports the single-model results.

\begin{table}[!htbp]
\centering
\stoplossmainstyle

\setlength{\tabcolsep}{4pt}
\renewcommand{\arraystretch}{1.12}
\renewcommand{\tabularxcolumn}[1]{m{#1}}

\begin{threeparttable}

\caption{
\textbf{Trained ensembles achieve higher failure-prediction AUC
than their untuned backbones in all four settings.}
Parentheses give absolute AUC gains over the corresponding
untuned base.
Step count only uses the current tool-step index as the failure
score, without reading trajectory content.
}
\label{tab:full}

\begin{tabularx}{\linewidth}{
    @{}
    >{\raggedright\arraybackslash}m{7.3em}
    >{\raggedright\arraybackslash}m{6.2em}
    *{4}{>{\centering\arraybackslash}X}
    @{}
}
\toprule
&
& \multicolumn{2}{c}{\textbf{GPT-5.6 Luna runs}}
& \multicolumn{2}{c}{\textbf{Claude Sonnet 4.6 runs}} \\
\cmidrule(lr){3-4}
\cmidrule(lr){5-6}
\textbf{Backbone}
& \textbf{Configuration}
& \shortstack{SWE-bench\\Verified}
& \shortstack{SWE-bench\\Pro}
& \shortstack{SWE-bench\\Verified}
& \shortstack{SWE-bench\\Pro} \\
\midrule
& Untuned base
& 0.491
& 0.452
& 0.512
& 0.511 \\
\rowcolor{stoplossgray}
\cellcolor{white}\multirow{-2}{=}{\textbf{Qwen2.5-7B-Instruct}}
& \shortstack[l]{\textbf{Trained}\\\textbf{ensemble}}
& \shortstack{0.622\\(+0.131)}
& \shortstack{0.614\\(+0.162)}
& \shortstack{0.561\\(+0.049)}
& \shortstack{0.579\\(+0.068)} \\
\addlinespace[0.65ex]
& Untuned base
& 0.504
& 0.507
& 0.481
& 0.516 \\
\rowcolor{stoplossgray}
\cellcolor{white}\multirow{-2}{=}{\textbf{Qwen3-8B}}
& \shortstack[l]{\textbf{Trained}\\\textbf{ensemble}}
& \shortstack{0.605\\(+0.100)}
& \shortstack{0.621\\(+0.114)}
& \shortstack{0.563\\(+0.082)}
& \shortstack{0.608\\(+0.092)} \\
\addlinespace[0.65ex]
& Untuned base
& 0.437
& 0.433
& 0.455
& 0.478 \\
\rowcolor{stoplossgray}
\cellcolor{white}\multirow{-2}{=}{\textbf{Qwen2.5-Coder-7B-Instruct}}
& \shortstack[l]{\textbf{Trained}\\\textbf{ensemble}}
& \shortstack{0.612\\(+0.175)}
& \shortstack{0.620\\(+0.187)}
& \shortstack{0.561\\(+0.106)}
& \shortstack{0.578\\(+0.100)} \\
\midrule
\multicolumn{2}{@{}l}{Step count only}
& 0.489
& 0.528
& 0.471
& 0.513 \\
\bottomrule
\end{tabularx}
\end{threeparttable}
\end{table}

\textbf{The trained monitors also outperform a simple step-count baseline.}
This baseline assigns higher failure scores to windows reached
after more tool steps, without examining what happened during execution.
Its tier AUC ranges from 0.471 to 0.528, below every trained
ensemble in the corresponding setting (Table~\ref{tab:full}).
The learned scores therefore rank failures more effectively
than tool-step count alone.

\textbf{The predictive signal is already present at the first scheduled check, tool step 5.}
The primary ensemble achieves ordinary AUCs of 0.642 on
Verified/Luna and 0.645 on Pro/Luna, compared with 0.494 and
0.450 for its untuned base.
These results show that training improves failure prediction
early in execution.
Figure~\ref{fig:checkpoint} reports performance at later checkpoints;
Section~\ref{sec:timing} examines when the stopping policy
actually acts.

\subsection{Prediction quality and monitoring cost}
\label{sec:paired}

StopLoss provides failure predictions at low monitoring cost. The primary ensemble's estimated inference cost is \$0.000237 per check under our GPU-time accounting. Under the tested
configurations, GPT judges cost 9--74 times as much per check, pooling each judge's checks across its two benchmarks. On GPT-5.6 Luna trajectories, the primary ensemble's tier AUC is within 0.044 of each judge's point estimate. This combination of low monitoring cost and predictive discrimination helps explain the policy-level cost advantages in Section~\ref{sec:deploy}. On Claude Sonnet~4.6 trajectories, Claude Opus~5 achieves higher AUC (Table~\ref{tab:judges}; Section~\ref{sec:transfer}). Appendix~\ref{app:accounting} details the cost accounting.

\section{Analysis and Discussion}
\label{sec:analysis}

\subsection{When are failures recognized and stopped?}
\label{sec:timing}

\begin{figure}[t]
\centering
\includegraphics[width=0.9\linewidth]{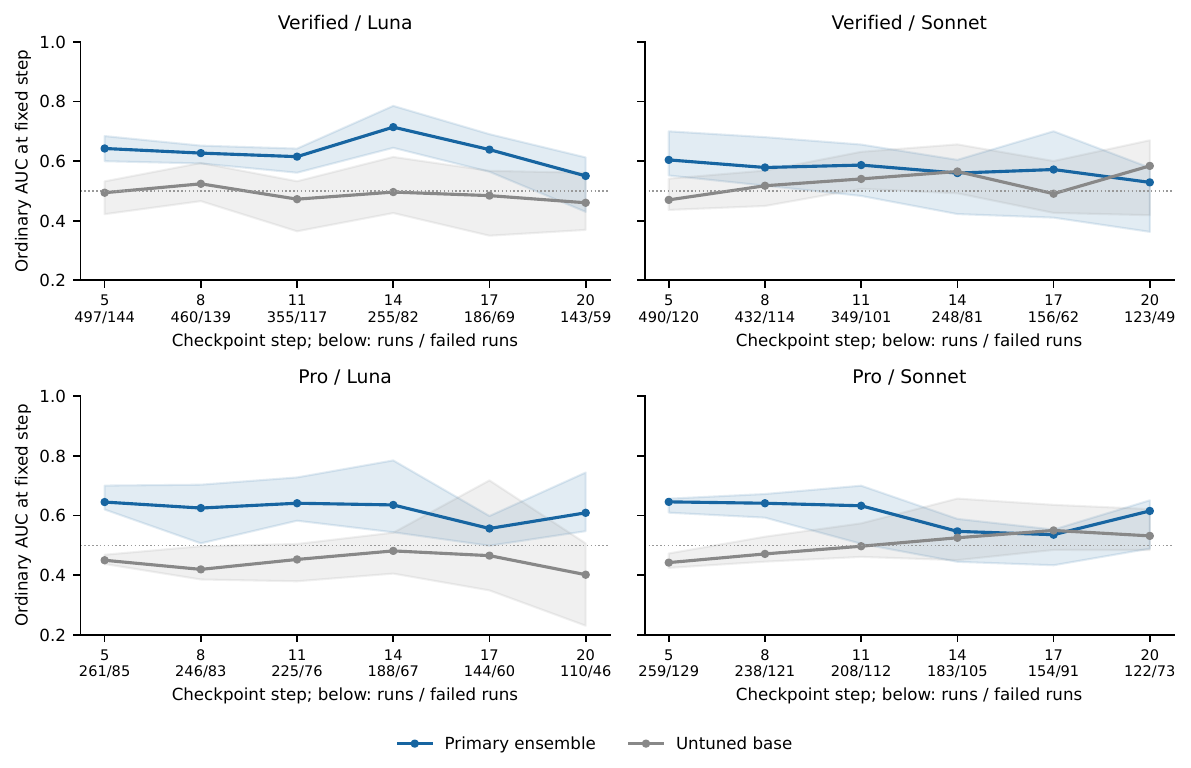}
\caption{
\textbf{StopLoss provides an early failure signal from the first scheduled check.}
At tool step 5, the primary StopLoss ensemble (blue) achieves higher
ordinary AUC than its untuned base (gray) in all four settings.
Higher AUC indicates better separation of eventual failures from
successes; the dotted line marks chance-level ranking (AUC = 0.5).
This post-hoc evaluation uses one observation per original run
reaching each checkpoint, without applying the stopping policy.
Shaded bands show pointwise 95\% repository-bootstrap intervals.
Numbers below each checkpoint report total runs / failed runs.
Later checkpoints contain different subsets of runs.
}
\label{fig:checkpoint}
\end{figure}

Figure~\ref{fig:checkpoint} extends the first-check results in
Section~\ref{sec:detect} across scheduled checkpoints.
On Claude Sonnet~4.6 runs, the primary monitor also shows predictive signal
at tool step 5, with ordinary AUCs of 0.604 on Verified
and 0.645 on Pro.
To examine stopping timing, we audit all Verified/Luna
runs under the frozen GPT-calibrated threshold.
Across the eight correctly stopped failures, the median fraction
of execution cost remaining when the policy triggered was 64.5\%.
The three largest savings account for 65.8\% of the total
execution cost avoided by these correct stops.
Thus, even a small number of correct stops can avoid substantial
remaining execution, with savings concentrated in a few runs.

\subsection{Transfer of predictions and stopping thresholds}
\label{sec:transfer}
\label{sec:calibration}

\looseness=-1
\textbf{StopLoss learns failure cues from local action--observation
windows, and these cues vary with how an executor uses
tools and responds to errors.}
A post-hoc audit shows that failed runs are longer than
successful runs under both executors on both benchmarks,
so the length--failure association also holds on Claude Sonnet~4.6.
However, even at the same checkpoint, tool step 8 on Verified,
the primary monitor achieves an ordinary AUC of 0.627 on
GPT-5.6 Luna versus 0.578 on Claude Sonnet~4.6, as shown in
Figure~\ref{fig:checkpoint}.
This fixed-checkpoint gap supports the interpretation that
failure cues learned from GPT trajectories transfer only
partially to Claude Sonnet~4.6.
Claude Sonnet~4.6 calibration adjusts the stopping threshold
but does not change the learned scores or their ranking.

\looseness=-1
\textbf{Similar prediction quality can also produce different
stopping behavior across benchmarks.}
On GPT-5.6 Luna runs, tier AUC is 0.622 on Verified and 0.614 on Pro,
yet the frozen threshold yields false-kill rates of
1.0\% and 27.8\%, respectively, on the independent stopping
cohorts in Table~\ref{tab:judges}.
StopLoss stops a run as soon as any checkpoint score
crosses the threshold.
The run-length audit shows that successful GPT-5.6 Luna runs
have a median of 16 tool steps on Pro versus 13 on Verified.
Longer trajectories allow more scheduled checks and may
therefore increase the chance of a threshold crossing.
This provides a plausible contributor to the higher
false-kill rate on Pro, alongside shifts in
the scores assigned to successful runs.
Threshold transfer thus depends on the scores encountered
throughout a successful trajectory, beyond the ranking
quality summarized by AUC.

\subsection{Execution cost and monitoring overhead}
\label{sec:costdrivers}

Let $C=\sum_t C_t(\infty)$, let $g$ be the fraction of execution cost avoided by aborts, and let $m=\sum_t M_t/C$ be monitoring cost relative to $C$. Writing $r=\mathrm{FK}$, a policy retaining at least one success satisfies
\begin{equation}
 \frac{\mathrm{CPS}}{\mathrm{CPS}_0}
 =\frac{1-g+m}{1-r},
 \qquad \mathrm{CPS}<\mathrm{CPS}_0\ \Longleftrightarrow\ g>m+r.
 \label{eq:break-even}
\end{equation}
Execution savings must cover both monitoring overhead and lost successes. A small monitor reduces $m$; whether its threshold yields enough savings $g$ at an acceptable false-kill rate $r$ remains empirical.

On Verified/Luna, the 425 complete runs cost \$28.44, while GPT-5.6 Sol's visited checks alone cost \$33.20 and StopLoss's cost \$0.44. This monitoring term explains why similar ranking points can produce very different costs per retained solve. On Verified/Sonnet, Claude Opus~5 avoids 16.0\% of execution cost but spends 44.9\% of the original execution budget on monitoring, raising cost per solve by 34.7\%. More expensive execution, as on Pro/Sonnet, leaves room for judges to recover their monitoring expense (see Table~\ref{tab:judges}).

\paragraph{Sensitivity to the inference-cost proxy.}
We measure cost-per-solve changes relative to full execution while inflating only the ensemble's monitoring charge, holding scores, thresholds, visited checks, and false-kill rates fixed. On the independent Verified/Luna stopping cohort, doubling the charge moves the point estimate from $-4.5\%$ to $-2.9\%$; at five times the charge it becomes $+1.8\%$. Both Pro cells retain savings at ten times the proxy, with unchanged stopping risks (see Appendix~\ref{app:costsensitivity}).

\section{Related Work}

\paragraph{Trajectory evaluation and early stopping.}
Reward models and verifiers assess reasoning and agent
trajectories to support verification, search, and corrective
feedback
\citep{lightman2024verify,xia2025agentrm,gandhi2025astray}.
Online failure monitoring turns trajectory assessment into
early intervention.
\citet{ruan2026doomed} use hidden-state probes with
recall-controlled stopping, while AgentStop relies on
inexpensive execution signals such as token log probabilities
\citep{pham2026agentstop}.
PrefixGuard learns failure-warning monitors from observable
traces \citep{huang2026prefixguard}.
FailFast--RestartSmart trains a small text-based monitor
using terminal outcomes and intermediate patch-test
supervision, then combines stopping with restarts that offer
previous edits as an optional overlay \citep{wang2026failfast}.
StopLoss learns eventual failure from final
verification outcomes and short action--observation windows,
providing inexpensive stopping decisions without requiring
executor internals or intermediate patch-test labels.

\paragraph{Cost-efficient agent execution.}
BAGEN studies remaining-resource estimation and completion
within a budget \citep{lin2026bagen}, while EET uses
execution experience to guide early termination during
patch generation and selection \citep{guo2026eet}.
FrugalGPT and RouteLLM reduce inference costs through
model cascades and routing
\citep{chen2023frugalgpt,ong2024routellm}.
SWE-Router uses partial coding trajectories to decide
whether to continue with a cheaper model or escalate
to a stronger one \citep{son2026swerouter}.
StopLoss focuses on whether to continue the current run
under a fixed executor, using predicted failure rather
than budget exhaustion as its stopping signal.
We evaluate this decision through cost per retained solve,
accounting for monitoring charges and lost successes,
and compare the trade-offs with prompted judges.


\section{Limitations and Conclusion}

\label{sec:limitations}

Our models are trained exclusively on GPT-generated trajectories, which may limit their generalization to Claude executions. Budget constraints also restrict our training and evaluation to Python tasks, with our SWE-bench Pro evaluation covering only three repositories from its public Python subset. We believe StopLoss provides a general approach to early stopping that can be extended to other programming languages and benchmarks using suitable trajectories and outcome labels. We leave this broader evaluation to future work.

StopLoss shows that a small monitor trained on outcomes can make early stopping pay for itself. It learns only from completed trajectories and their verifier outcomes, and it stops likely failures during execution without changing the executor. Outcome supervision lifts failure prediction from near chance to a useful signal by the first scheduled check, and each check costs one to two orders of magnitude less than a call to a general-purpose judge. In all four executor--benchmark settings and with all three backbones, StopLoss has lower cost per retained solve than every judge policy. It saves 4.5\% on Verified/Luna while keeping 99\% of successful runs, and up to 31\% on Pro (48\% with a \mbox{Qwen3-8B} backbone). Our results also show that ranking failures well is not enough: the execution that a monitor saves must exceed both its own cost and the successes it stops. Improving cross-executor generalization and threshold transfer is the next step toward bringing this inexpensive monitoring to more agent settings.

\subsection*{AI Use Statement}
We used generative AI tools to assist with language polishing and textual revisions of the manuscript, including improvements to clarity, grammar, and presentation. The authors take responsibility for the final content of the paper.

\subsection*{Ethics statement}
The experiments use public software repositories and benchmark tasks. We preserve their licensing requirements. An early-stop policy can silently reduce an agent's success rate; we report false kills and retained solves alongside resource savings to make this trade-off explicit.

\bibliography{iclr2027_conference}
\bibliographystyle{iclr2027_conference}
\clearpage
\appendix
\raggedbottom

\section{Prompts, labels, and training details}
\label{app:prompt}
\label{app:recipes}

The boxes reproduce the user-message template and judge suffix, with placeholders and line breaks normalized for typesetting. Each window contains up to three tool-step action/observation blocks in order. Construction first truncates each action (reasoning plus commands) to 700 characters and each observation to 900 characters. If the resulting monitor sequence exceeds 1,600 tokens, the builder removes text from the middle of the window body, retaining its beginning and end and marking the omission. It preserves the step-count header and monitor question, checking the full chat-templated prompt plus every candidate answer. The trainer and scorer then reject any overlong sequence; they do not silently truncate it again.

\begin{promptbox}{StopLoss monitor: user-message template}
\small
An autonomous agent has run $\langle k\rangle$ steps on a task. Its last steps:

\texttt{<action>}\\
$\langle$action text$\rangle$\\
\texttt{</action>}\\
\texttt{<observation>}\\
$\langle$observation text$\rangle$\\
\texttt{</observation>}

$\langle$repeat the block for the remaining recent steps$\rangle$

Decide the agent's best move now: CONTINUE working, DELIVER (verify and
declare done), or ABORT (hopeless). Answer with one word.
\end{promptbox}

The score is the softmax over the first answer tokens, read as the probability of ABORT. For \mbox{Qwen3-8B} the chat template has thinking disabled. Weak outcome labels do not certify that a run could not recover under a different continuation.

Paid judges receive exactly the same window body, including any truncation, so every method sees the same evidence. Only the final question, shown below, and each provider's chat formatting differ, because a prompted judge must be asked for a probability, whereas the small monitor reads it from its answer logits. The 1,600-token limit above does not restrict judge outputs or reasoning.

\begin{promptbox}{General-purpose judge: replacement question}
\small
You are an expert reviewer of autonomous coding-agent runs. The agent above
is working on a software-engineering task in a repository; only its most
recent steps are shown. Estimate the probability that this run will
ULTIMATELY FAIL (its final patch will not pass the hidden tests) if the agent
keeps going as it has been. Reply with at most three sentences of reasoning,
then a final line exactly of the form:\\
\texttt{FAIL\_PROBABILITY: <integer 0-100>}
\end{promptbox}

The parsed probability is divided by 100. Score-acceptance checks are specified in Appendix~\ref{app:pitfalls}. Configurations and prompts remain frozen throughout the full-coverage evaluation.

\paragraph{Executor and judge configurations.}
Table~\ref{tab:generation-config} separates trajectory generation from judging, including when the same model serves both roles. Same-model self-judging follows the stated judging configuration, which need not match the executor's generation configuration. This distinction does not change the fixed input windows or model-family restriction. The trial configurations and benchmark task files for all 1,523 valid runs record 100 agent turns and separate 7,200-second agent and verifier timeouts. The collector also has a 16,000-second outer process timeout. These are execution limits; monitor checkpoints count tool steps. The recorded scaffold is Terminus-2 version 2.0.0 in Harbor 0.17.0.

\begin{table}[H]
\centering
\caption{Recorded generation and judging configurations. Model identifiers are listed in Table~\ref{tab:prices}. Collection spend valves use the collection-time accounting schedule; reported costs are recomputed under the frozen full-price schedule.}
\label{tab:generation-config}
\small
\setlength{\tabcolsep}{4pt}
\renewcommand{\arraystretch}{1.12}
\begin{tabularx}{\linewidth}{@{}>{\raggedright\arraybackslash}p{0.15\linewidth}>{\raggedright\arraybackslash}p{0.19\linewidth}>{\raggedright\arraybackslash}p{0.24\linewidth}>{\raggedright\arraybackslash}X@{}}
\toprule
Role & Model(s) & Reasoning setting & Limits and interface \\
\midrule
Executor & GPT-5.6 Luna & Effort unset; upstream default not recorded &
100 turns; 7,200 s agent timeout; \$9 spend valve. Output-token caps stripped by the GPT gateway. \\
\addlinespace
Executor & Claude Sonnet~4.6 & Thinking enabled, 8,000-token budget &
100 turns; 7,200 s agent timeout; \$26 spend valve. Gateway default output limit 16,000 tokens. \\
\addlinespace
Judge & GPT-5.6 Sol,\newline GPT-5.6 Terra,\newline GPT-5.6 Luna & High reasoning effort &
Native Responses API; output-token caps stripped, so no enforced per-call output cap. \\
\addlinespace
Judge & Claude Opus~5,\newline Claude Opus~4.6,\newline Claude Sonnet~4.6 & Adaptive thinking, high effort &
Native Messages API; output limits 4,000, 8,000, 16,000 on truncation retries. \\
\bottomrule
\end{tabularx}
\end{table}

\paragraph{Encoding and training objective.}
The corpus and task split are specified in Appendix~\ref{app:datasplits}. A training input is the chat-templated prompt followed by its complete target word, with no appended EOS token. The prompt and right-padding labels are masked with $-100$, and every answer token is supervised by the standard causal language-model loss over the vocabulary. The main recipe uses zero label smoothing and does not enable the alternative three-logit soft-target trainer. Each candidate word spans two tokens under the three base tokenizers. Scoring uses only the first-token readout in Section~\ref{sec:scoring}.

The shared encoder checks that the prompt/answer token boundary aligns. The label boundary remains $k\geq n_t-1$ for successful runs, independently of the sampled window grid.

\paragraph{Optimization and selection.}
All three bases use one epoch, rank-16 LoRA with scale 32 and dropout 0.05 on q/k/v/o attention projections, AdamW with learning rate $10^{-4}$ and weight decay 0.05, cosine decay, and 3\% warm-up. The effective batch size is 16: Qwen2.5-7B-Instruct uses microbatch 2 with accumulation 8 on one device; Qwen2.5-Coder-7B-Instruct and Qwen3-8B use two devices with microbatch 2 and accumulation 4 per device. Validation occurs every 300 optimizer steps and at the end of training, selecting the checkpoint with best failure AUC on the fixed validation checkpoint windows. Scoring uses the audited BF16 autocast and FP32 readout softmax for both calibration and test.

\section{Data and evaluation protocol}
\label{app:extension}

\subsection{Data construction and splits}
\label{app:datasplits}

The original SWE-rebench corpus contains 5,086 tasks and 63,173 windows; 3,986 tasks have \mbox{GPT-5.6} Luna trajectories and 1,100 have \mbox{GPT-5.6} Terra trajectories. Corpus windows originate from the grid $\{2,3,4,5,6,8,10,12,15,20,25,30,40,50,60,75,100\}$, with additional all-step windows in later collections; evaluation uses policy checkpoints $5,8,11,\ldots$. We reserve 219 tasks from held-out repositories, comprising 1,858 windows, for model selection. Before optimization, the trainer removes every window belonging to those tasks, leaving 4,867 fitting tasks and 61,315 windows. The external Verified and Pro tasks are disjoint from both partitions.

For external evaluation, we retain all collected runs with valid verdicts for ranking. We construct the independent stopping cohort by task identity: remove the 122 early exploratory tasks (74 Verified and 48 Pro), then retain only tasks with valid verdicts from both executors. The resulting shared task lists contain 425 Verified and 213 Pro tasks. They exclude all Claude Sonnet~4.6 calibration tasks; GPT calibration tasks belong to the separate SWE-rebench validation partition. Section~\ref{sec:cohorts} explains which reported results use each cohort.

\subsection{Threshold calibration}
\label{app:thresholds}

\paragraph{Eligible runs and checkpoint windows.}
From the successful-run pools specified in Section~\ref{sec:cohorts}, we retain runs reaching at least one policy checkpoint. The GPT all-step rendering yields 348 checkpoint windows from 90 successful runs. Of the 76 successful Claude Sonnet~4.6 runs, 74 reach a checkpoint and yield 378 windows; the remaining two do not enter threshold computation. All methods evaluating the same executor family use the same calibration runs and windows.

\paragraph{Threshold computation.}
For each method, take the maximum checkpoint score on each of the $n$ eligible successful calibration runs, and sort these maxima as $z_{(1)}\geq\cdots\geq z_{(n)}$. With target rate $\rho$, set $b=\lfloor\rho n\rfloor$ and $\tau=z_{(b+1)}+10^{-9}$. The offset places the threshold strictly above the selected maximum because stopping uses $s_t(k)\geq\tau$. At most $b$ calibration successes are stopped; tied maxima are not split and can make the rule more conservative.

\subsection{Evaluation coverage}
\label{app:coverage}

Table~\ref{tab:coverage-detail} adds solved-run counts, runs without checkpoints, and checkpoint-window counts to the cohort definitions above. Infrastructure or verifier failures without a verdict are excluded, not labeled as model failures; executor solve-rate denominators remain the original 500 Verified and 266 Pro targets. Runs without checkpoints remain in cohort counts and stopping denominators but contribute no ranking windows or monitor calls. Full-cohort stopping replays, including the timing diagnostic in Section~\ref{sec:timing}, are descriptive; the main policy comparisons use the independent stopping cohort.

\begin{table}[H]
\centering
\caption{Detailed evaluation coverage. Solved runs, runs without checkpoints, and checkpoint windows refer to the full ranking cohort. Every checkpoint window is scored by all three same-family judges and all small monitors. The last column gives the calibration-disjoint stopping cohort.}
\label{tab:coverage-detail}
\footnotesize
\setlength{\tabcolsep}{4pt}
\resizebox{\linewidth}{!}{%
\begin{tabular}{lrrrrr}
\toprule
Benchmark / executor & Ranking runs & Solved & No checkpoint & Windows & Stopping runs \\
\midrule
Verified / Luna & 499 & 355 & 2 & 2307 & 425 \\
Verified / Sonnet & 499 & 377 & 9 & 2215 & 425 \\
Pro / Luna & 264 & 177 & 3 & 1569 & 213 \\
Pro / Sonnet & 261 & 132 & 2 & 1960 & 213 \\
\bottomrule
\end{tabular}}
\end{table}

\paragraph{Coverage extension.}
The full-coverage experiment is a separately recorded extension decided after inspecting the initial judge results and full small-monitor scores; it was not fixed by the initial registration. Throughout the extension, we held the primary monitor, model configurations, window definition, calibration thresholds, prices, and comparison rules fixed. The extension collected 16,029 new valid scores and reused 8,124 accepted scores, including eligible Claude calibration windows, after checking inputs and configurations; accepted windows were not scored again. All 24,153 required judge--window scores are present, covering 3,876 windows per GPT judge and 4,175 per Claude judge. The manuscript reports this full-coverage analysis without reclassifying it as the original fixed-sample experiment.

\paragraph{Score acceptance and provenance checks.}
\label{app:pitfalls}
Task identity is resolved from the trial configuration. Before analysis, verifier labels, score records, prompt and configuration hashes, and call ledgers are checked for consistency. Accepted judge scores require successful parsing, normal completion, usage records, and an allowed backend model identity. Truncated answers are rejected even if they contain a parsable number. These checks bind measurements to their recorded inputs and calls.

\section{Cost accounting and sensitivity analysis}
\label{app:accounting}
\label{app:replay}

\subsection{Token pricing and execution accounting}
\label{app:tokenpricing}

\begin{table}[H]
\centering
\caption{Frozen September 18 full-price schedule in dollars per million tokens. Every input token is charged at the input rate, including cache reads and writes; output includes reasoning/thinking tokens.}
\label{tab:prices}
\begin{tabular}{lrr}
\toprule
Model identifier & Input & Output \\
\midrule
\texttt{gpt-5.6-sol} & 4.00 & 20.00 \\
\texttt{gpt-5.6-terra} & 2.00 & 12.00 \\
\texttt{gpt-5.6-luna} & 0.20 & 1.20 \\
\texttt{claude-opus-5} & 5.00 & 25.00 \\
\texttt{claude-opus-4-6} & 5.00 & 25.00 \\
\texttt{claude-sonnet-4-6} & 3.00 & 15.00 \\
\bottomrule
\end{tabular}
\end{table}

This is a standardized cost comparison under the frozen schedule, not an estimate of discounted or subscription charges. The source configuration records the provider pricing-page checks and rejects main-panel calls above its short-context guard rather than silently selecting a tariff. Tool-prefix spend is aligned to ledger calls; surviving runs also pay trailing non-tool calls. Abort decisions avoid future execution and future checks only.

\subsection{Monitor inference cost}
\label{app:monitorcost}

The score metadata records NVIDIA RTX A6000 GPUs, batch size 1, and BF16 autocast with FP32 readout softmax. The proxy divides elapsed scoring-loop wall time by completed windows at the last progress record (27,001 windows). For the five Qwen2.5-7B-Instruct adapters, this is 0.211--0.215 seconds per window; summing members gives 1.065 GPU-seconds and \$0.000237 at \$0.80/GPU-hour. One adapter averages about \$0.000047 per check. The loop includes tensor creation, forward computation, and probability transfer to the host, but begins after model loading and prompt tokenization and ends before the final consistency pass and output writing. This is an amortized scoring proxy, not measured online request latency or a latency percentile. It is reused across evaluation cells; concurrent ensemble execution, serving overhead, and utilization were not benchmarked. Training, fixed deployment, and idle-resource costs are outside this proxy.

Across all trained seeds, mean single-model / ensemble costs per check are \$0.0000473 / \$0.0002367 for Qwen2.5-7B-Instruct, \$0.0000572 / \$0.0002289 for Qwen3-8B, and \$0.0000473 / \$0.0001893 for Qwen2.5-Coder-7B-Instruct. Single-model costs average the members; ensemble costs sum them, using the same GPU-time proxy.

\begin{table}[H]
\centering
\caption{Mean per-prediction judge costs on the same full-cohort windows, using recorded attempts at full token prices. Ratios divide by the primary ensemble proxy of \$0.000237 per check. This is independent of which checks a stopping policy visits.}
\label{tab:unitcost}
\footnotesize
\setlength{\tabcolsep}{4pt}
\resizebox{\linewidth}{!}{%
\begin{tabular}{llrrr}
\toprule
Cell & Judge & Windows & \$/check & Judge / ensemble \\
\midrule
Verified / Luna & GPT-5.6 Sol & 2307 & 0.01949 & 82.3 \\
Verified / Luna & GPT-5.6 Terra & 2307 & 0.00774 & 32.7 \\
Verified / Luna & GPT-5.6 Luna & 2307 & 0.00248 & 10.5 \\
Verified / Sonnet & Claude Opus 5 & 2215 & 0.07722 & 326.3 \\
Verified / Sonnet & Claude Opus 4.6 & 2215 & 0.02015 & 85.1 \\
Verified / Sonnet & Claude Sonnet 4.6 & 2215 & 0.02114 & 89.3 \\
Pro / Luna & GPT-5.6 Sol & 1569 & 0.01446 & 61.1 \\
Pro / Luna & GPT-5.6 Terra & 1569 & 0.00618 & 26.1 \\
Pro / Luna & GPT-5.6 Luna & 1569 & 0.00182 & 7.7 \\
Pro / Sonnet & Claude Opus 5 & 1960 & 0.02594 & 109.6 \\
Pro / Sonnet & Claude Opus 4.6 & 1960 & 0.01696 & 71.7 \\
Pro / Sonnet & Claude Sonnet 4.6 & 1960 & 0.01634 & 69.0 \\
\bottomrule
\end{tabular}}
\end{table}

Across cells, judge costs range from \$0.0018 to \$0.0772 per prediction, approximately 8--326 times the primary ensemble proxy. Pooling each judge across its two benchmarks by window count gives ratios of 9--225; the GPT-only range reported in Section~\ref{sec:paired} is 9--74.

\subsection{Collection accounting}
\label{app:collection}

Table~\ref{tab:collection} reports the experimental bill for newly issued scoring calls in the full-coverage extension. This differs from the monitoring charge in policy CPS: offline evaluation covers all required windows, whereas the stopping replay charges only checks visited before stopping or completion. One-off calibration is accounted for separately below.

\begin{table}[H]
\centering
\caption{Collection accounting for the full-coverage extension. Covered windows include reused valid scores; new windows were scored once successfully, with any recorded failed attempts charged separately. Known dollars are the newly collected calls at list prices, excluding reused historical scores and calibration expenditure. Unknown-usage calls are additional failed attempts whose prices cannot be reconstructed; they are not priced at zero. The total is rounded after summing unrounded call costs.}
\label{tab:collection}
\footnotesize
\setlength{\tabcolsep}{4pt}
\resizebox{\linewidth}{!}{%
\begin{tabular}{lrrrr}
\toprule
Judge & Covered windows & New windows & Known new \$ & Unknown usage \\
\midrule
GPT-5.6 Sol & 3876 & 2751 & 48.64 & 3 \\
GPT-5.6 Terra & 3876 & 2751 & 19.71 & 2 \\
GPT-5.6 Luna & 3876 & 2751 & 6.12 & 7 \\
Claude Opus 5 & 4175 & 2592 & 146.36 & 47 \\
Claude Opus 4.6 & 4175 & 2592 & 49.07 & 0 \\
Claude Sonnet 4.6 & 4175 & 2592 & 49.68 & 1 \\
\midrule
Total & 24153 & 16029 & 319.57 & 60 \\
\bottomrule
\end{tabular}}
\end{table}

The extension adds 16,029 valid window scores with known recorded costs of \$319.57 at full list prices. It also contains 60 failed attempts with unavailable usage; there are 64 such recorded attempts across all source packages, including calibration. Their cost is unknown, not zero. At most two additional in-flight requests may have ended without ledger records. Valid scores cover every required window, but complete accepted scores do not imply that all failed requests have reconstructable charges. One-off calibration costs are \$5.67, \$2.34, and \$0.72 for GPT-5.6 Sol, GPT-5.6 Terra, and GPT-5.6 Luna, and \$20.14, \$6.95, and \$7.12 for Claude Opus~5, Claude Opus~4.6, and Claude Sonnet~4.6; these are separate from the new-call total and the policy replay bill.

\subsection{Sensitivity to monitoring cost}
\label{app:costsensitivity}

This post-hoc analysis multiplies the primary ensemble's monitoring charge by $\lambda\in\{1,2,5,10\}$, holding predictions, thresholds, visited checkpoints, stopping positions, and retained solves fixed. Executor and judge prices also remain fixed. Let $E$ be the execution cost incurred under the policy, $M$ its original monitoring charge, and $S$ the number of retained solves. Then
\[
 \mathrm{CPS}_{\lambda}=\frac{E+\lambda M}{S}.
\]
At $\lambda=1$, all four settings on the stopping cohort reproduce the frozen analysis point estimates. Table~\ref{tab:cost-sensitivity-deployment} reports the adjusted cost-per-solve changes; no confidence intervals are recomputed.

\begin{table}[H]
\centering
\caption{Stopping-cohort cost-per-solve change under inflated primary-ensemble monitoring charges. Scores, frozen nominal-5\% thresholds, visited checks, and false-kill rates remain fixed. Negative values are savings.}
\label{tab:cost-sensitivity-deployment}
\footnotesize
\setlength{\tabcolsep}{4pt}
\begin{tabular*}{\linewidth}{@{\extracolsep{\fill}}lrrrrr@{}}
\toprule
Cell & $1\times$ & $2\times$ & $5\times$ & $10\times$ & FK\% \\
\midrule
Verified / Luna & -4.5 & -2.9 & +1.8 & +9.7 & 1.0 \\
Verified / Sonnet & +0.2 & +0.3 & +0.8 & +1.5 & 0.0 \\
Pro / Luna & -24.2 & -23.0 & -19.5 & -13.6 & 27.8 \\
Pro / Sonnet & -30.6 & -30.5 & -30.3 & -29.9 & 8.4 \\
\bottomrule
\end{tabular*}
\end{table}

Verified/Luna's cost-per-solve saving disappears when the monitoring charge reaches approximately $3.83\times$ its baseline value. Both Pro settings retain savings at $10\times$, with success retention unchanged. Verified/Sonnet stops no runs and therefore only adds monitoring cost.

\clearpage
\section{Prediction and stopping diagnostics}
\label{app:diagnostics}

\begin{table}[H]
\centering
\stoplossmainstyle
\begin{threeparttable}
\caption{\textbf{Complete training-seed results.} Single-model tier AUC of every trained seed on the full ranking cohort.}
\label{tab:seeds-full}
\begin{tabularx}{\linewidth}{l>{\hsize=1\hsize\linewidth=\hsize\centering\arraybackslash}X>{\hsize=1\hsize\linewidth=\hsize\centering\arraybackslash}X>{\hsize=1\hsize\linewidth=\hsize\centering\arraybackslash}X>{\hsize=1\hsize\linewidth=\hsize\centering\arraybackslash}X}
\toprule
\multirow{2}{*}{\textbf{Seed}} & \multicolumn{2}{c}{\textbf{GPT-5.6 Luna runs}} & \multicolumn{2}{c}{\textbf{Claude Sonnet 4.6 runs}} \\
\cmidrule(lr){2-3}\cmidrule(lr){4-5}
 & \multicolumn{1}{c}{Verified} & \multicolumn{1}{c}{Pro} & \multicolumn{1}{c}{Verified} & \multicolumn{1}{c}{Pro} \\
\midrule
\rowcolor{stoplossscenariogray}
\multicolumn{5}{l}{\textbf{Qwen2.5-7B-Instruct}} \\
Seed 0 & 0.615 & 0.607 & 0.578 & 0.548 \\
Seed 1 & 0.608 & 0.599 & 0.522 & 0.564 \\
Seed 2 & 0.605 & 0.586 & 0.531 & 0.586 \\
Seed 3 & 0.598 & 0.610 & 0.545 & 0.582 \\
Seed 4 & 0.607 & 0.599 & 0.566 & 0.568 \\
\rowcolor{stoplossgray}
\textbf{Mean $\pm$ SD} & 0.606 $\pm$ 0.006 & 0.600 $\pm$ 0.010 & 0.549 $\pm$ 0.023 & 0.569 $\pm$ 0.015 \\
\addlinespace[0.65ex]
\rowcolor{stoplossscenariogray}
\multicolumn{5}{l}{\textbf{Qwen3-8B}} \\
Seed 0 & 0.602 & 0.621 & 0.549 & 0.610 \\
Seed 1 & 0.604 & 0.608 & 0.540 & 0.594 \\
Seed 2 & 0.602 & 0.614 & 0.566 & 0.601 \\
Seed 3 & 0.604 & 0.616 & 0.576 & 0.598 \\
\rowcolor{stoplossgray}
\textbf{Mean $\pm$ SD} & 0.603 $\pm$ 0.001 & 0.615 $\pm$ 0.005 & 0.558 $\pm$ 0.016 & 0.601 $\pm$ 0.007 \\
\addlinespace[0.65ex]
\rowcolor{stoplossscenariogray}
\multicolumn{5}{l}{\textbf{Qwen2.5-Coder-7B-Instruct}} \\
Seed 0 & 0.614 & 0.617 & 0.556 & 0.578 \\
Seed 1 & 0.598 & 0.610 & 0.502 & 0.571 \\
Seed 2 & 0.587 & 0.616 & 0.551 & 0.595 \\
Seed 3 & 0.616 & 0.604 & 0.593 & 0.536 \\
\rowcolor{stoplossgray}
\textbf{Mean $\pm$ SD} & 0.604 $\pm$ 0.013 & 0.612 $\pm$ 0.006 & 0.550 $\pm$ 0.037 & 0.570 $\pm$ 0.025 \\
\bottomrule
\end{tabularx}
\end{threeparttable}
\end{table}

\begin{samepage}
\paragraph{Illustrative stopping cases.}
For each outcome category in the full Verified/Luna cohort, we select the middle run ordered by complete execution cost, breaking ties by task identity. All three cases use the same threshold, 0.876, and illustrate behavior rather than error prevalence.
\begin{itemize}
\item \textbf{Correct stop: SymPy-12419.} At step 17, the window shows continued investigation of nested symbolic summation and Piecewise conditions. The failure score is 0.894, and the logged continuation fails verification. Stopping avoids \$0.225 of the run's \$0.306 execution cost.
\item \textbf{Missed failure: SymPy-20428.} The maximum checkpoint score is 0.550 at step 11. The window reports passing local tests and readiness to finish, but the completed run fails verification.
\item \textbf{False kill: scikit-learn-25931.} At step 20, repeated terminal-recovery attempts and echoed keystrokes receive a failure score of 0.955, although the completed run passes verification.
\end{itemize}
\end{samepage}

\end{document}